\documentclass{article}

\usepackage{arxiv}

\usepackage[utf8]{inputenc} 
\usepackage[T1]{fontenc}    
\usepackage{hyperref}       
\usepackage{url}            
\usepackage{booktabs}       
\usepackage{amsfonts}       
\usepackage{nicefrac}       
\usepackage{microtype}      
\usepackage{lipsum}
\usepackage{graphicx}

\usepackage{times}
\usepackage{epsfig}

\usepackage{amsmath}
\usepackage{amssymb}

\usepackage{subfigure}
\usepackage{algorithm}
\usepackage{algpseudocode}
\usepackage{mathtools} %
\usepackage{dsfont} %
\usepackage{bbm}
\usepackage{booktabs} %
\usepackage{diagbox} 
\usepackage{nccmath} %
\usepackage{arydshln} %
\usepackage{tablefootnote}
\usepackage{bbding} %
\usepackage{enumitem} %

\usepackage{color}
\usepackage{wrapfig}
\usepackage{multirow}
\begin{document}

\title{Conterfactual Generative Zero-Shot Semantic Segmentation}

\author{Feihong Shen \\
Jilin University \\
shenfh5517@mails.jlu.edu.cn
        \And
        Jun Liu \\
Singapore University of Technology and Design \\
junliu@sutd.edu.sg
        \And
        Ping Hu \\
        Boston University\\
pinghu@bu.edu
        }







\maketitle
\begin{abstract}
Zero-shot learning is an essential part of computer vision. As a classical downstream task, zero-shot semantic segmentation has been studied because of its applicant value. One of the popular zero-shot semantic segmentation methods is based on the generative model Most new proposed works added structures on the same architecture to enhance this model. However, we found that, from the view of causal inference, the result of the original model has been influenced by spurious statistical relationships. Thus the performance of the prediction shows severe bias. In this work, we consider counterfactual methods to avoid the confounder in the original model. Based on this method, we proposed a new framework for zero-shot semantic segmentation. Our model is compared with baseline models on two real-world datasets, Pascal-VOC and Pascal-Context. The experiment results show proposed models can surpass previous confounded models and can still make use of additional structures to improve the performance. We also design a simple structure based on Graph Convolutional Networks (GCN) in this work.
\end{abstract}


\section{Introduction}
In some real-world applications, we know some classes (seen class) and we want to get the segmentation of other classes (unseen class) we have never seen before.  This task, what we called zero-shot semantic segmentation, requires mature mechanisms to extract word embeddings and visual features. Luckily, developed CNN-based~\cite{chen2017deeplab,chen2018encoder} and FCN-based~\cite{long2015fully}  make it accessible to get features from images. Word2vec~\cite{mikolov2013distributed} and Fast-test~\cite{joulin2016fasttext} can help us transfer class categories to word embeddings. In the other word, what we need to do is make use of both visual and text knowledge to pixel-levelly classify images which contain never-seen objects before.

Previous works~\cite{zhang2015zero,socher2013zero,xian2018feature} have already revealed some rational ways to use this knowledge and build up a zero-shot model. We generally use three groups of methods to train unsupervised representation: transformation-equivariant representations~\cite{hinton2011transforming}, self-supervised methods~\cite{jing2020self} and generative models~\cite{goodfellow2014generative}. In the spectrum of unsupervised methods, zero-shot learning always tries to get the visual knowledge of unseen classes by learning the mapping from word embedding to visual features~\cite{fu2018recent}. As a zero-shot learning task, some researchers also took the afore-mentioned groups of methods to handle zero-shot semantic segmentation, especially generative models.
\begin{figure}
    \centering
    \includegraphics[width=0.95\linewidth]{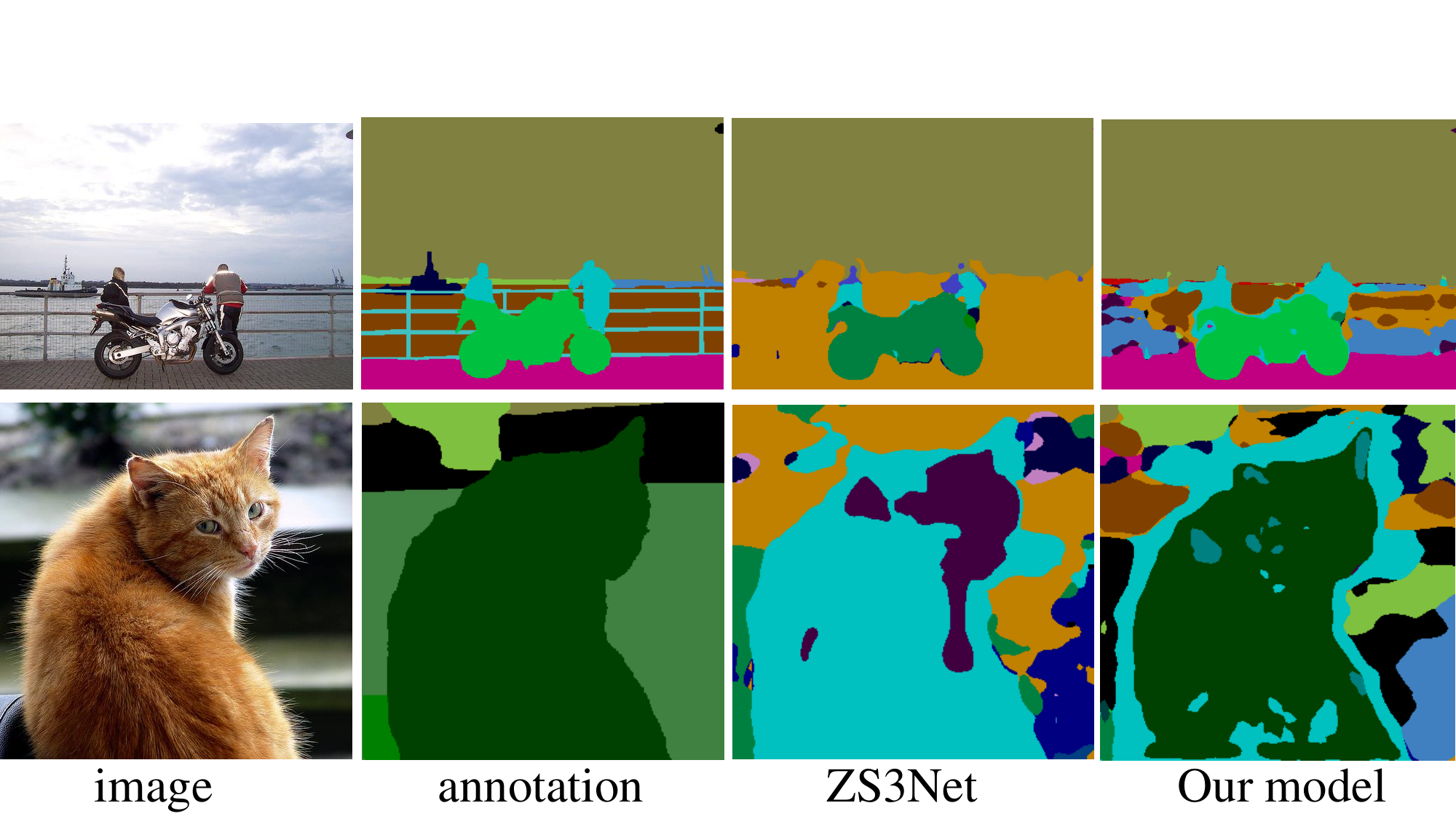}
    \caption{The first column is the original images that contain unseen classes: motorbike and cat. The second column is the pixel-level ground truth (GT) of the semantic segmentation. In the traditional generative zero-shot semantic segmentation model ZS3Net, the objects were incorrectly classified as their similar classes: bicycle and dog. These two classes are visible in the training set and this dataset bias was alleviated in our model..}
    \label{fig:example}
\end{figure}
We mainly focus on generative zero-shot semantic segmentation~\cite{lv2020learning,gu2020context,tian2020cap2seg} in this work. While we could get flamboyant accuracy of zero-shot prediction with powerful computing resources and various training tricks, the semantic correlations between word embeddings and visual features in generative models may not be correctly explored. The most common manifestation of this error is that the output of the model is based on a statistical relationship rather than causal relationships between the parameters. Previous causal inference work has demonstrated the shortcoming of statistical relationship~\cite{pearl2009causal}. Specifically, in the computer vision field, the defect of pure statistical relationship may worsen dataset bias~\cite{tang2020long}. In zero-shot semantic segmentation, this bias represents intuitively by the imbalance of the accuracy between seen classes and unseen classes~\cite{hu2020uncertainty}. We aim to alleviate this imbalance in this work as shown in Figure~\ref{fig:example}.

Correspondingly, there are also many causal methods to solve causal problems in computer vision models. To our best known, backdoor and front door adjustment has been used to handle spurious correlation in few-shot learning and image caption~\cite{yue2020interventional,yang2020deconfounded}; direct and indirect effect help to generate unbiased scene graph~\cite{tang2020unbiased}; counterfactual causality was widely used in visual question answer (VQA) problem and long-tailed classification~\cite{yu2020counterfactual,niu2020counterfactual}.

In this work, we aim to alleviate bias between seen class and unseen class in zero-shot semantic segmentation. To achieve this goal, we proposed an easy-to-use training strategy based on causality and apply it to previous models. Our modified models surpass their original version on the same datasets. 

Our contribution can be summarized as follows: (1) Inspired by previous causal inference in the computer vision field, we proposed a new strategy to reduce the unbalance in zero-shot semantic segmentation. Our strategy can be used on all generative models in this task and develop their performance. (2) Since our strategy is based on causal inference, we can explain why recent different structures of models can develop the performance of traditional work. (3) Following the explanation, we extend the model with our structures and improve performance.

\section{Related Work}

\noindent \textbf{zero-shot semantic segmentation.} While the term of zero-shot semantic segmentation is from earlier work, the recent development of this task is based on two works: ZS3Net~\cite{bucher2019zero} and SPNet~\cite{xian2019semantic}. They also represent two categories of models in this task: embedding and generative. ZS3Net is a traditional generative model that generates fake features through word embeddings of unseen classes. SPNet generates multi-modal knowledge concating visual features and word embeddings. Most of the new works follow these two works and import new knowledge resources like context~\cite{gu2020context,bucher2019zero} and relationships between classes~\cite{li2020consistent} to help generate more fitting fake features. To some point, these new knowledge can reduce the imbalance between seen and unseen classes. We will introduce the reason in the following sections. However, different from previous works, we fundamentally change the training methods and keep the integrity of original models.

\noindent \textbf{counterfactual computer vision.} As an important direction in causal inference, counterfactual analysis has inspired many computer vision works~\cite{goyal2019counterfactual,yu2020counterfactual,niu2020counterfactual}. In the factual world, parameters in a single system influence the value of each other through the mechanism of the system. However, in the counterfactual world, we can block special links of this mechanism and quantitatively analyze the interaction of these parameters. The casual implementation of the counterfactual world in computer vision tasks is commonly the modification of the image. In this work, we apply counterfactual analysis to figure out the indirect effect of visible features on the final prediction. One Very recent paper also leverages the counterfactual method to figure out the bias in generative zero-shot learning~\cite{yue2021counterfactual}, our work is also different from that work.

\section{Method}\label{sec:method}

\begin{figure*}
    \centering
    \includegraphics[width=0.95\linewidth]{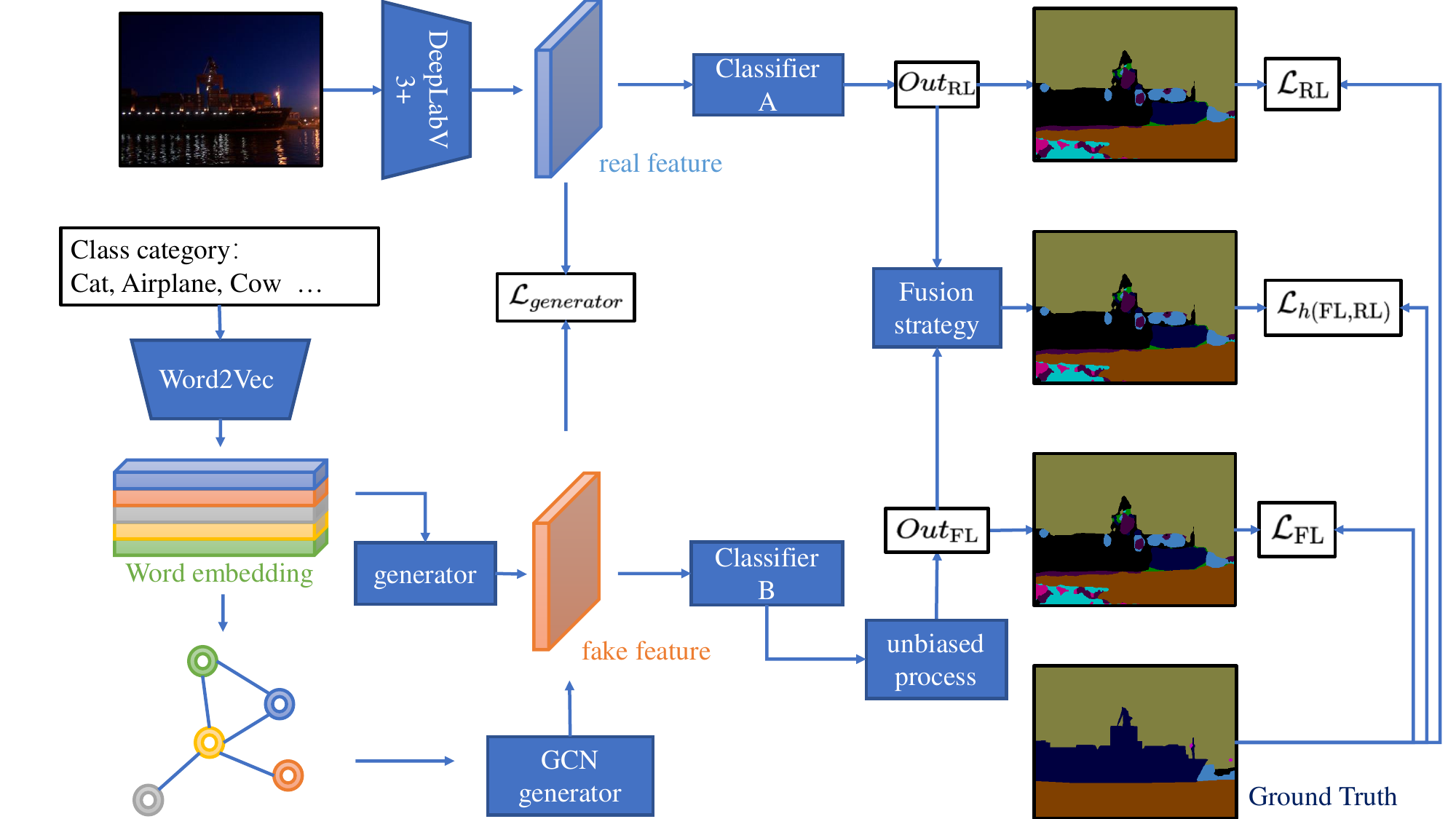}

    \caption{Illustration of our counterfactual generative zero-shot semantic segmentation framework. In the training phase, once one image input, the backbone extracts the visible features. We take these features to train our classifier A and generators. If there are unseen classes in the annotation of this picture, we use fake features from generators to train our classifier B. We apply the deconfounder process to the output of classifier B. The output of these two classifiers is mixed up by our fusion strategy. In the testing phase, we directly cast features into classifier A and B. The output of the fusion function will be our final prediction.} 

    \label{fig:overview}
\end{figure*}
\subsection{Problem Definitions}
\begin{figure}
\centering
{
\subfigure[Factual model]{
\label{fig:2-causal-graph-a}
\includegraphics[width=0.45\linewidth]{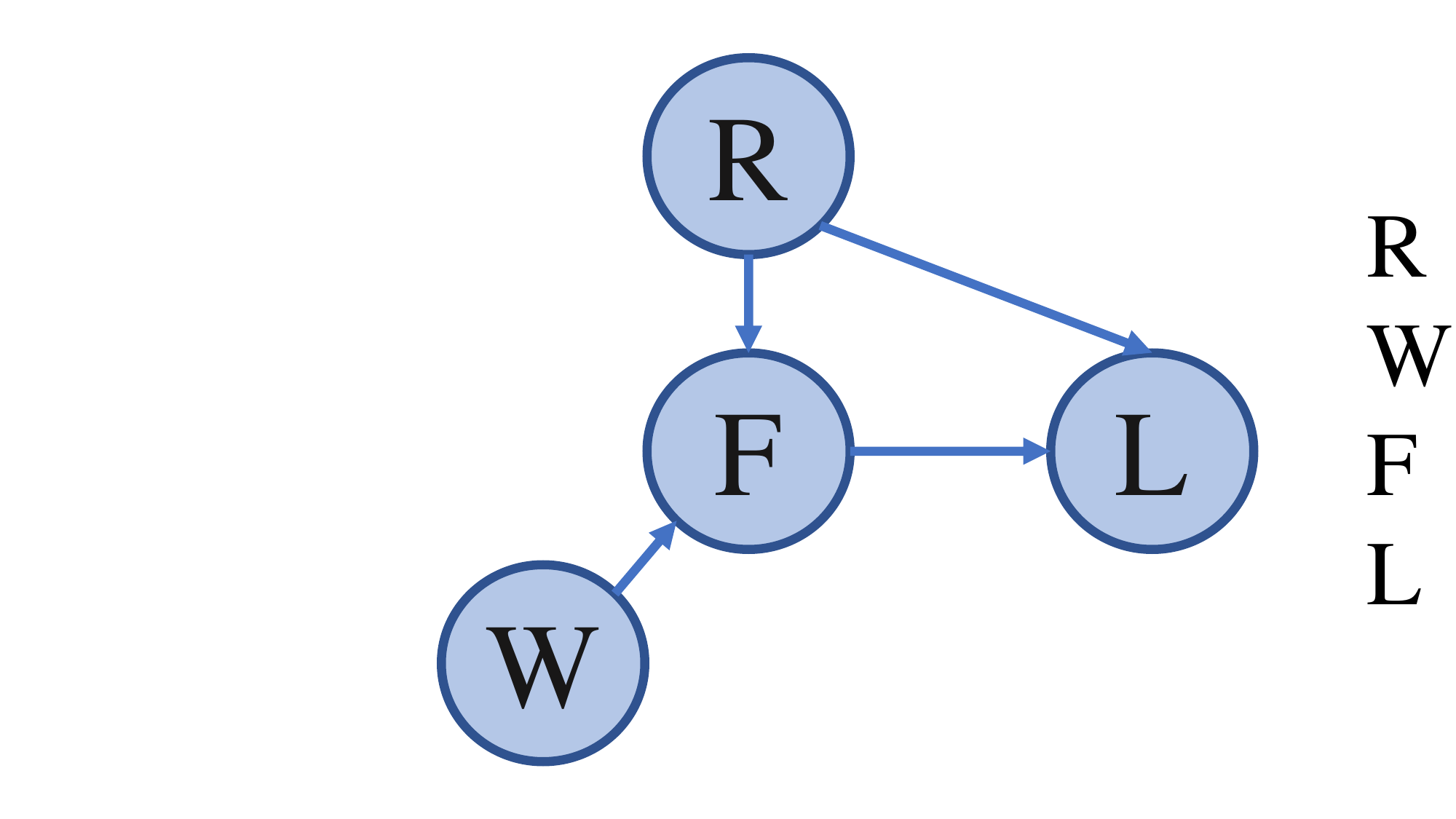}
}
\subfigure[Counterfactual model]{
\label{fig:2-causal-graph-b}
\includegraphics[width=0.45\linewidth]{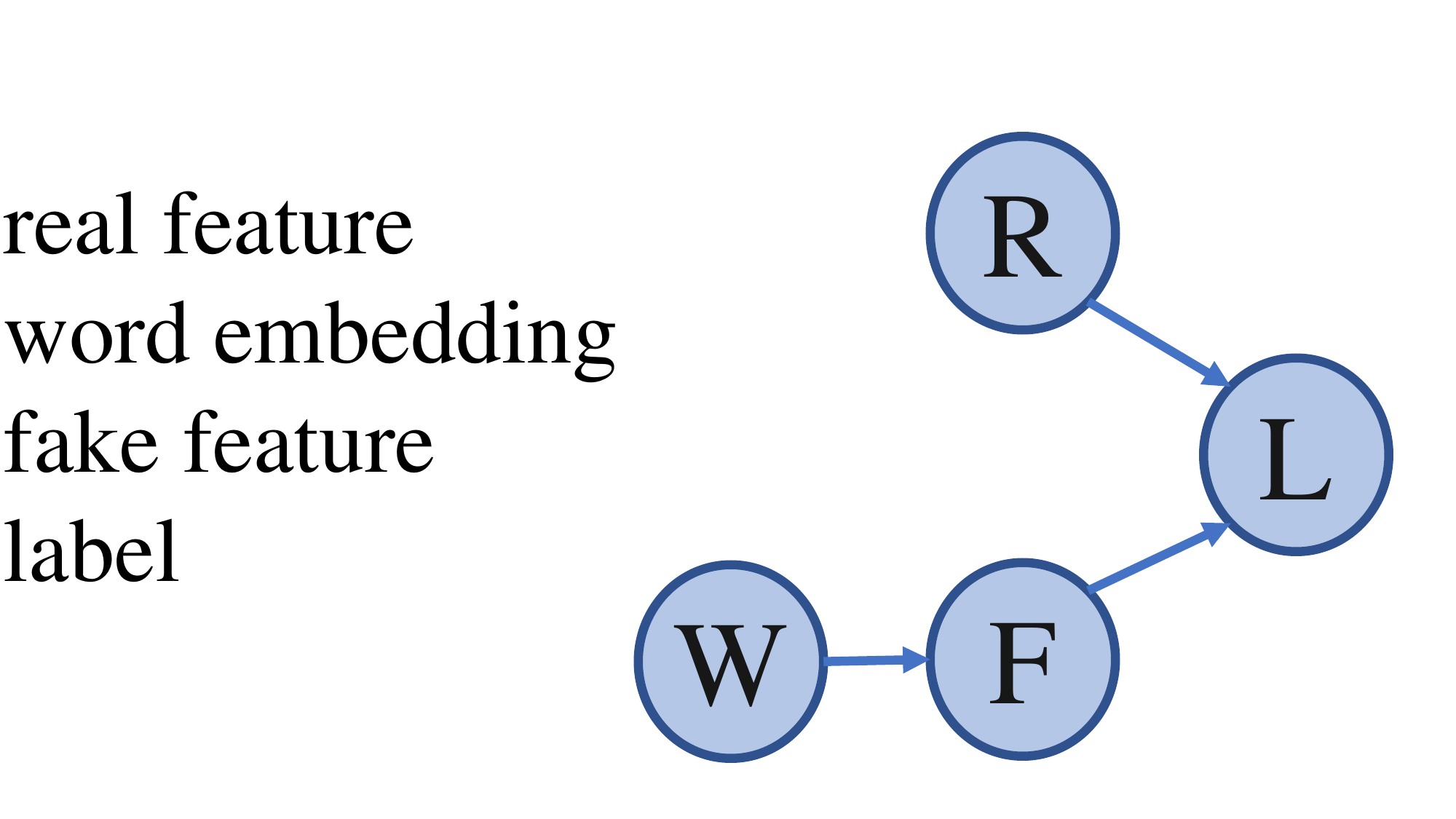}
}
\subfigure[model of SPNet]{
\label{fig:spnet}
\includegraphics[width=0.45\linewidth]{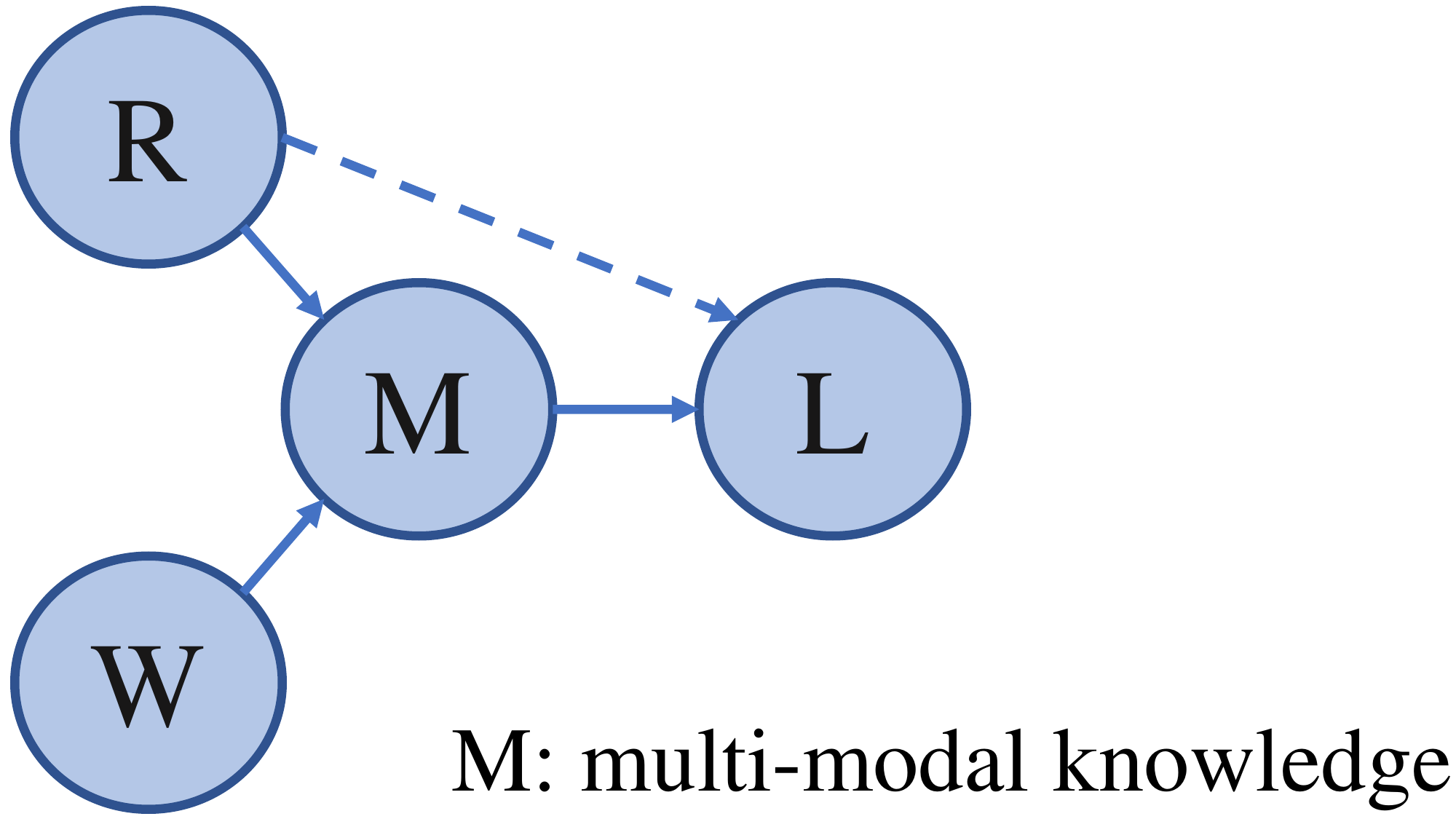}
}
}

\caption{(a)The causal structure of traditional generative zero-shot semantic segmentation. (b)The ideal deconfounder model for zero-shot semantic segmentation (c)The causal structure of SPNet}
\label{fig:2-causal-graph}
\end{figure}

We define the problem by describing the causal relationships between the variables of one traditional generative zero-shot segmentation model in Figure~\ref{fig:2-causal-graph-a}. Node $R$ denotes the ``real'' features extracted by the backbone from the images, corresponding to node $F$: fake features generated by the generator. What ``fake feature” means is the object which this feature represents is invisible in the training set but present in the testing set. This is also the difference between zero-shot semantic segmentation and traditional semantic segmentation.  Node W denotes the word embedding of objects in images and node $L$ is the image-level labels produced by classifiers. The directed links can be simply understood as the causal effect between variables. Now we detail the rationale behind these links high-levelly.

$R \rightarrow L$. Real feature $R$ can directly affect the label $L$. In the traditional model, the feature extracted by up-stream architectures is independently cast into a convolution layer to train the final classifier. 

$F \rightarrow L$. Since in zero-shot learning, samples of the unseen class are missing, the model uses a generator to get the fake feature with the same format as real features. Similarly to $R \rightarrow L$, the fake features also get involved in the training of the final classifier, which determines every pixel-level prediction. 

$R \rightarrow F\leftarrow W$. Word embedding $W$ includes both seen class and unseen class. The generator learns on seen features and word embedding of seen class to generate features of one object from its word embedding. These two variables determined the way that word embedding of unseen class generates fake features $F$. 

The above mechanism seems very logical and rational from a statistical view. However, this traditional model cannot capture the pure effect of real features on the label because the real features R not only determine the label $L$ by the link $R \rightarrow L$ but also indirectly influence the label by path $R \rightarrow F \rightarrow L$. This structure, a.k.a. confounder, misleading the effect of fake feature F to the label L and eventually cause dataset bias in zero-shot learning. For instance, the prediction of classifiers tends to collapse to the motorbike rather than the bicycle when the image of the bicycle is unseen in the dataset. In causality, There are two ways to alleviate this situation: one is importing other variables (e.g. context) to strengthen fake features and minimize the percentage of effect real features are taking. It is also the common method that previous work takes. In causality, we want mechanisms in the system not to change or give information to any of the other mechanisms. This is called Independent Causal Mechanism (ICM) Principle~\cite{scholkopf2021toward}. In this zero-shot task, the basic mechanisms are $R \rightarrow L$ and $W \rightarrow F \rightarrow L$. The other way is following is the principle by removing the indirect effect of real features to label and then get the new intervened causal graph like Figure~\ref{fig:2-causal-graph-b}.

It seems like the structure of the SPNet in Figure~\ref{fig:spnet} can also apply the independent causal mechanisms principle. In spite of the fact that this method reaches great accuracy in some specific situation, we regard generative zero-shot semantic segmentation as a more explainable and potential method: it is hard to explain what knowledge M represent in feature space and auxiliary knowledge, e.g. context, cannot directly help to enhance the SPNet-like structure.
\subsection{Indirect Effect Removing}
Before removing the indirect effect, we allocate real features a separate classifier to keep the causal relationship $R \rightarrow L$. We call this branch $R-L$ model   After we get an unbiased model for path $W \rightarrow F \rightarrow L$, fusion strategy will be applied to the outputs of these two models.

While in statistical we formulate a conditional probability form:
\begin{equation}
    \mathcal{P} \{ L = l  | W = w, R = r, F = f( W = w, R = r)\}
\end{equation}
to represent the prediction of the model, we replace the upper formulation with:
\begin{equation}
    L_{w, r, f_{w,r}}=L(W=w, R=r, F=f)
\end{equation}

With different treatments of parameters $W=w^{*}, R=r^{*}, F=f^{*}$ , the value of prediction is: 
\begin{equation}
    L_{w^{*}, r^{*}, f^{*}_{w^{*}, r^{*}}}=L(W=w^{*}, R=r^{*}, F=f^{*})
\end{equation}
. In the factual world, the value of $F$ is determined by $W$ and $R$, however, in the counterfactual world, the prediction can be:
\begin{equation}
    L_{w^{*}, r^{*}, f_{w,r}}=L(W=w^{*}, R=r^{*}, F=f)
\end{equation}
We can get the total effect of W and R on L following the previous work about causal effect:
\begin{equation}
    TE = L_{w, r, f_{w,r}} - L_{w^{*}, r^{*}, f^{*}_{w^{*}, r^{*}}},
\end{equation}
Then we should remove the direct effect of real feature $R$ on the $L$, since we have another branch to represent causal effect $F \rightarrow L$. In order to get the natural direct effect $(NDE)$ of real feature $R$ on label $L$, we block the link $R \rightarrow F$. In the counterfactual world, the $NDE$ is the difference between a blocked and an unblocked causal graph.
\begin{equation}
    NDE = L_{w^{*}, r, f^{*}_{w^{*}, r^{*}}} - L_{w^{*}, r^{*}, f^{*}_{w^{*}, r^{*}}},
\end{equation}
As we mentioned before, the bias of unseen class mainly comes from the indirect effect of real features through path $R \rightarrow F \rightarrow L$. While we still want to keep the advantage of generating procedure, the counterfactual method still can mathematically minimize the natural indirect effect $(NIE)$
\begin{equation}
    NIE = L_{w^{*}, r^{*}, f_{w^{*}, r}} - L_{w^{*}, r^{*}, f^{*}_{w^{*},r^{*}}},
\end{equation}
We called the branch of this unbiased path $W \rightarrow F \rightarrow L$ as $F-L$ model. We use  $Out_{\text{RL}}$ to denote the output of the $R-L$ model and $Out_{\text{FL}}$ to denote the output of the $F-L$ model. So the $Out_{\text{FL}}$ can be calculated by:
\begin{equation}\label{eq:out}
\begin{split}
    Out_{FL} &= TE-NDE-NIE \\
    &= L_{w, r, f_{w,r}} - L_{w^{*}, r, f^{*}_{w^{*}, r^{*}}}- L_{w^{*}, r^{*}, f_{w^{*}, r}}+ L_{w^{*}, r^{*}, f^{*}_{w^{*}, r^{*}}}
\end{split}
\end{equation}
Remember that the $NDE$ and $NIE$ are all the effect from the real feature $R$ but the $TE$ is from both $R$ and $W$. So here
\begin{equation}
    TE \neq NDE+NIE
\end{equation}

\subsection{Counterfactual Implement}
In the implement of the whole architecture, we use no-treatment as the reference value of R and W (i.e. $R\!=\!r^*\!=\!\varnothing$ and $W\!=\!w^*\!=\!\varnothing$) Then we can define the value of the effect:
\begin{equation}
\begin{split}
    L =
    \begin{cases}
    L_{w, r, f_{w,r}} & \text{ if $W=w$ and $R=r$}\\
	a L_{w, r, f_{w,r}}(a=0,1) & \text{ if $W=\varnothing$ and $R=r$}\\
    c & \text{ if $W=\varnothing$ and $R=\varnothing$}\\
    \end{cases},
\end{split}
\end{equation}
We use a constant $c$ to represent the random guess of the network while the input is void. This method is widely used in previous counterfactual work. We could still get the trained model without the word embedding input $W$ through the classifier will lose the ability to recognize the unseen class in images. That is also the reason why we multiply $a$ on the $L_{w, r, f_{w,r}}$. As a result, we can get the final representation of $Out_{\text{FL}}$:
\begin{equation}\label{eq:fl}
    Out_{FL} = (1-a) L_{w, r, f_{w,r}}-(c_{1}+c_{2})
\end{equation}
That being said, to get the unbiased output of path $W\rightarrow F \rightarrow L$, we just need to multiply a constant on the output of the traditional model. In the real experiment, we use a trainable parameter to replace $a$ for convenience.

The final prediction of the model should be the fusion of $Out_{\text{RL}}$ and $Out_{\text{FL}}$. It is obvious that the calculation of the effect is based on the assumption that the whole effect system is linear, which requires the fusion strategy to keep consistency. There exists a difference between the contribution of real feature $F$ and word embedding $W$, so we cannot directly add up $Out_{\text{RL}}$ and $Out_{\text{FL}}$. Weighted sum-up might be a good idea but we already have the hyperparameter $a$ and do not want two more trainable parameters to break the interpretability of our model. 

Inspired by the kalman filter, we take use of the variant of the feature to balance the $Out_{\text{RL}}$ and $Out_{\text{FL}}$. We assume that $R$ and $F$ is i.i.d. and both apply the gaussian distribution, so the fusion result of the architecture will be:
\begin{flalign}
\mkern4mu h(Out_{\text{FL}},Out_{\text{RL}})=\frac{var(R)\cdot Out_{\text{FL}}+ var(F)\cdot Out_{\text{RL}}}{var(R)+var(F)}
\end{flalign}
where $h(\cdot)$ denotes fusion function and $var(\cdot)$ means the calculation of variation. So we optimized our model by minimize the sum of following losses:
\begin{equation}
    \mathcal{L}_{pred}=\mathcal{L}(Out_{\text{FL}})+\mathcal{L}(Out_{\text{RL}})+\mathcal{L}(h(Out_{\text{FL}},Out_{\text{RL}}))
\end{equation}
And $\mathcal{L}(\cdot)$ is the loss function regarding notations of the image as the targets.

\begin{figure}
    \centering
    \includegraphics[width=0.95\linewidth,trim={0cm 10cm 0cm 0cm},clip]{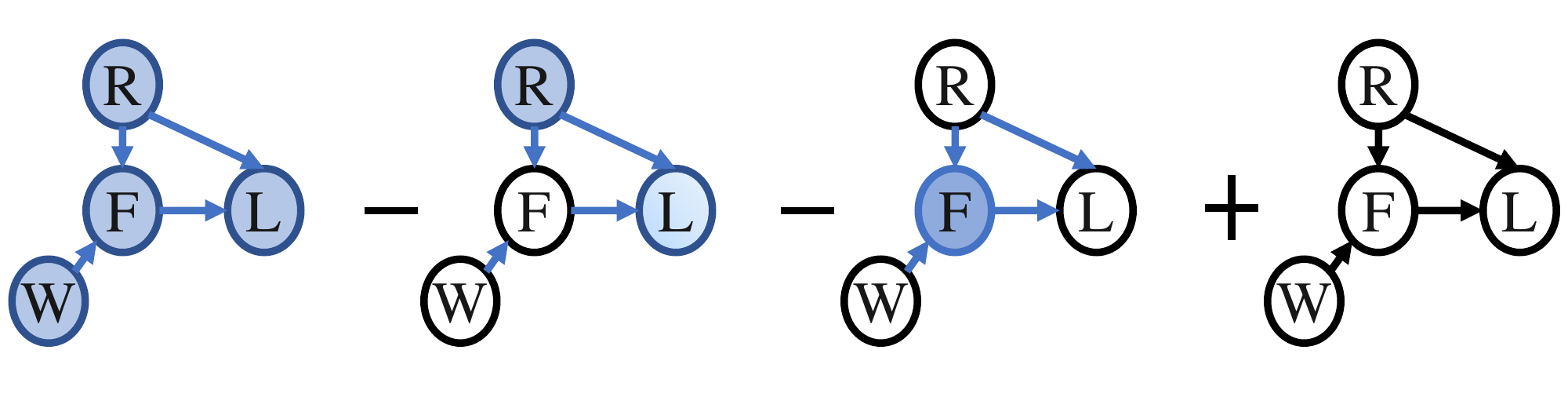}
    \vspace{-2mm}
    \caption{Our cause-effect look at Eq.~\ref{eq:out}. One of our model's training branches is based on this equation.}
    \vspace{-3mm}
    \label{fig:teaser}
\end{figure}

\begin{table*}[t]
\centering

\scalebox{0.81}{
\begin{tabular}{l c c c cccc c cccc}
\hline
\toprule
Dataset & & & & \multicolumn{4}{c}{Pascal-Context} & & \multicolumn{4}{c}{Pascal-Voc}\\  
\cmidrule{5-8} \cmidrule{10-13}
Model & ~ & Base. & ~ & {hIou} & {overall mIou} &  {Seen mIou} & {Unseen mIou} & ~ & {hIou} & {overall mIou} &  {Seen mIou} & {Unseen mIou}\\
\midrule
&&&&\multicolumn{4}{l}{\it two unseen classes: cow/motorbike}
&&\multicolumn{4}{l}{\it two unseen classes: cow/motorbike} \\
\hline
SPNet  &  & Embedding & &  31.77 & 37.06 & 37.38  & 27.63 & & 33.59 & 66.65   & 71.35  & 21.97 \\
CF-SPNet  &  & Embedding & &   30.82 & 36.81 & 37.18  & 26.32 & & 29.04 & {\bf 66.74} & {\bf 71.85} & 18.2 \\
ZS3Net  &  & Generative & & 7.96 & 27.70 & 28.50 & 4.63 & & 43.59 & 68.44 & 72.36 & 31.19 \\
CF-ZS3Net & & Generative & &  {\bf 33.07} & {\bf 39.13} & {\bf 39.49} & {\bf 28.46} & & 39.91 & 67.54 & 71.74 & 27.65 \\

\hline
&&&&\multicolumn{4}{l}{\it four unseen classes: upper+sofa/cat}&&\multicolumn{4}{l}{\it four unseen classes: upper+airplane/sofa} \\
\hline
SPNet  &  & Embedding & &  30.91 & 33.02 & 33.32  & 28.84 & & 30.6 & 60.14 & 69.67 & 19.61 \\
CF-SPNet  &  & Embedding & &   {\bf 31.24} & {\bf 33.36} & {\bf 33.66}  & {\bf 29.15} & & 25.31 & 52.69 & 61.33 & 15.95 \\
ZS3Net  &  & Generative & & 3.22 & 20.53 & 21.87 & 1.74 & & 35.44 & 60.33 & 68.91 & 23.85 \\
CF-ZS3Net & & Generative & &  {\bf 25.68} & {\bf 32.73} & {\bf 33.58} & {\bf 20.79} & & 35.29 & 58.95 & 67.19 & {\bf 23.93} \\

\hline
&&&&\multicolumn{4}{l}{\it six unseen classes: upper+boat/fence}&&\multicolumn{4}{l}{\it six unseen classes: upper+cat/tv} \\
\hline
SPNet  &  & Embedding & &  31.46 & 30.93 & 30.79  & 32.17 & & 21.62 & 46.64   & 60.02  & 13.19 \\
CF-SPNet  &  & Embedding & &   {\bf 33.38} & {\bf 32.27} & {\bf 31.98}  & {\bf 34.92} & & 21.16 & {\bf 46.82} & {\bf 60.41} & 12.83 \\
ZS3Net  &  & Generative & & 2.22 & 12.32 & 13.56 & 1.24 & & 30.43 & 40.63 & 47.97 & 22.29 \\
CF-ZS3Net & & Generative & &  {\bf 23.61} & {\bf 31.18} & {\bf 32.59} & {\bf 18.51} & & 29.35 & 35.88 & 41.1 & {\bf 22.83} \\

\hline
&&&&\multicolumn{4}{l}{\it eight unseen classes: upper+bird/tvmonitor}&&\multicolumn{4}{l}{\it eight unseen classes: upper+train/bottle} \\
\hline
SPNet  &  & Embedding & &  25.21 & 25.29 & 25.32  & 25.11 & & 20.82 & 38.00   & 53.43  & 12.93 \\
CF-SPNet  &  & Embedding & &   {\bf 26.65} & {\bf 26.38} & {\bf 26.29}  & {\bf 27.02} & & {\bf 22.32} & {\bf 38.62} & {\bf 53.71} & {\bf 14.09} \\
ZS3Net  &  & Generative & & 6.14 & 9.39 & 10.16 & 4.40 & & 24.85 & 26.57 & 29.79 & 21.33 \\
CF-ZS3Net & & Generative & &  {\bf 18.90} & {\bf 25.93} & {\bf 27.71} & {\bf 14.35} & & {\bf 30.82} & {\bf 46.1} & {\bf 61.83} & 20.53 \\

\hline
&&&&\multicolumn{4}{l}{\it ten unseen classes: upper+keyboard/aeroplane}&&\multicolumn{4}{l}{\it ten unseen classes: upper+chair/potted-plant} \\
\hline
SPNet  &  & Embedding & &  23.56 & 25.54 & 26.40  & 21.28 & & 21.34 & 35.30   & 55.36  & 13.22 \\
CF-SPNet  &  & Embedding & &   {\bf 24.26} & {\bf 26.63} & {\bf 27.63}  & {\bf 21.63} & & {\bf 22.75} & {\bf 35.92} & {\bf 55.56} & {\bf 14.31} \\
ZS3Net  &  & Generative & & 5.70 & 9.89 & 11.10 & 3.84 & & 23.28 & 26.23 & 33.97 & 17.71 \\
CF-ZS3Net & & Generative & &  {\bf 11.73} & {\bf 26.25} & {\bf 30.03} & {\bf 7.36} & & {\bf 30.49} & {\bf 43.55} & {\bf 65.04} & {\bf 19.92} \\
\midrule

\bottomrule
\end{tabular}
}
\vspace{2mm}
\caption{\textbf{Comparison (mIoU, hIoU) on Pascal-VOC 2012 and Pascal Context dataset.} \textbf{CF-} denotes our counterfactual training strategy. We highlighted the \textbf{results} that use our training strategy and are better than the performance of original models. The overall mIoU is the average pixel intersection-over-union (IoU) across all 21 and 59 classes and the unseen and seen mIoU only focuses on part of classes. The calculation of hIoU is in Eq.~\ref{eq:hiou}. We list the base of the zero-shot model in every row.  }
\label{tab:all}
\end{table*}

\subsection{GCN Structure}
After getting the deconfounder model, we still can take advantage of previous works about strengthening the link $W \rightarrow F$. While former researchers designed complicated structures to import context knowledge, on the basis of unbiased models, we only apply a simple graph convolution network (GCN) which is widely used in zero-shot learning~\cite{wang2018zero}. 

The generation of our fake features benefits from more available knowledge sources. In previous work~\cite{bucher2019zero}, the generators trained to generate fake features through the word embedding of a certain class while we believe that there exist many differences between the generation of ``keyboard’’ and ``sky’’ from their word embeddings for instance. We want our fake features to be able to learn from their similar classes and this requires the information simultaneously passing through the irregular data structure whose connection represents the relationship between classes. So we chose GCN as our fundamental building block.

We briefly introduce the preliminary in our GCN structure before describing our proposed framework.

We use a graph $\mathcal{G}=\{\mathcal{V},\mathcal{E}\}$ to represent relationships between classes, where $\mathcal{V}$ is the vertices of the graph representing the set of classes and each $\mathcal{E}$ is a weighted edge of the graph denoting similarity between these classes. Suppose we have $n$ classes of objects and $n_{unseen}$ classes of them is invisible in the image. We assume the format of pixel-level feature is $F \in R^{1\times\ell}$ and word embedding of class $n_1$ is $W_{n_{1}} \in \mathbb{R}^{1 \times w}$. $M\in \mathbb{R}^{n \times l}$ denotes an all-zero matrix and $A\in \mathbb{R}^{n \times n}$ denotes the adjacent matrix of the graph. 

\begin{figure}
    \centering
    \includegraphics[width=0.95\linewidth]{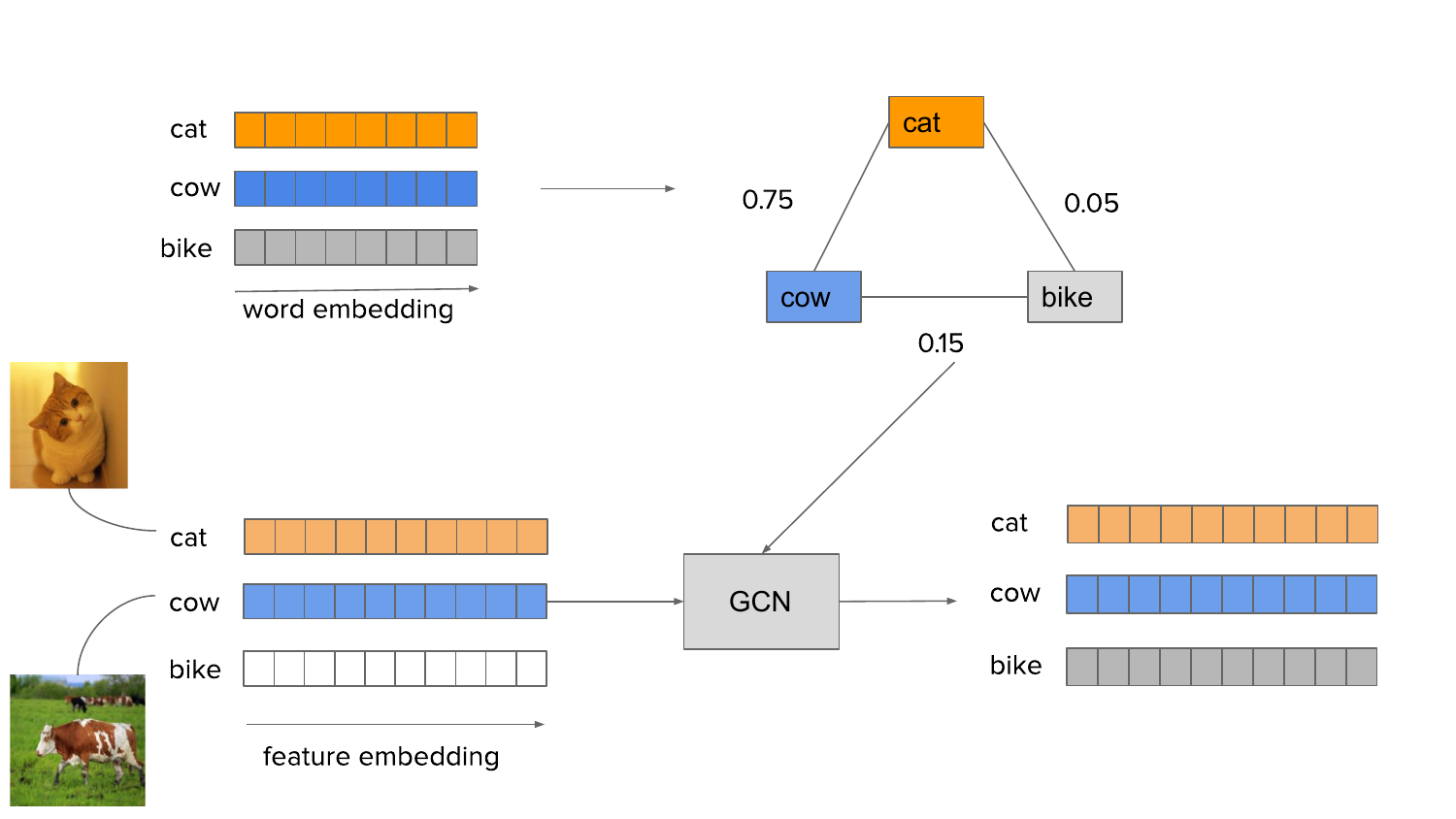}
    \caption{Suppose we have three classes of objects in our dataset: cat, cow, and bike. The images of the bike are unseen to our structure. After getting seen features from the backbone, we put them to their lines in the matrix and keep the line that represents the feature of the bike empty. After the graph convolution, we get the fake features of the bike class. In our structure, we regard the graph convolution network as a tool of message passing in an irregular data structure at a time and we get the message of unseen features from seen features in the relationship graph.} 
    \label{fig:gcn}
\end{figure}

Firstly, we use the word embeddings of the class to build the graph. We can calculate cosine similarity between each pair of word embeddings and the corresponding value of the adjacent matrix is changed if the value of the similarity surpasses a threshold $p$.

\begin{equation}
    A = \{ A_{n_{i},n_{j}} = c \; | \; cosine(W_{n_{i}},W_{n_{j}}) = c\;>\; p, n_{i,j}\in \mathcal{V}\}
\end{equation}
In the training period, each time the backbone extract feature $R$ of the class $n$, we clone the matrix $M$ in different memory and use this copy to train our generator. If the degree of the matrix $M$ reaches $n-n_{\text{unseen}}$, the graph convolutional layers perform convolution on this copy with our weighted adjacency matrix. Then we get the $n$ of the output as the feature of class $n$. We calculate the loss function between $R$ and this feature to train the GCN. After this process, the $n$ line of matrix $M$ will be replaced as $R$. Then we can get features of $n_{\text{unseen}}$ classes from matrix $M$. Figure~\ref{fig:gcn} shows our structure in a simple case.

\begin{figure*}
    \centering
    \includegraphics[width=0.95\linewidth]{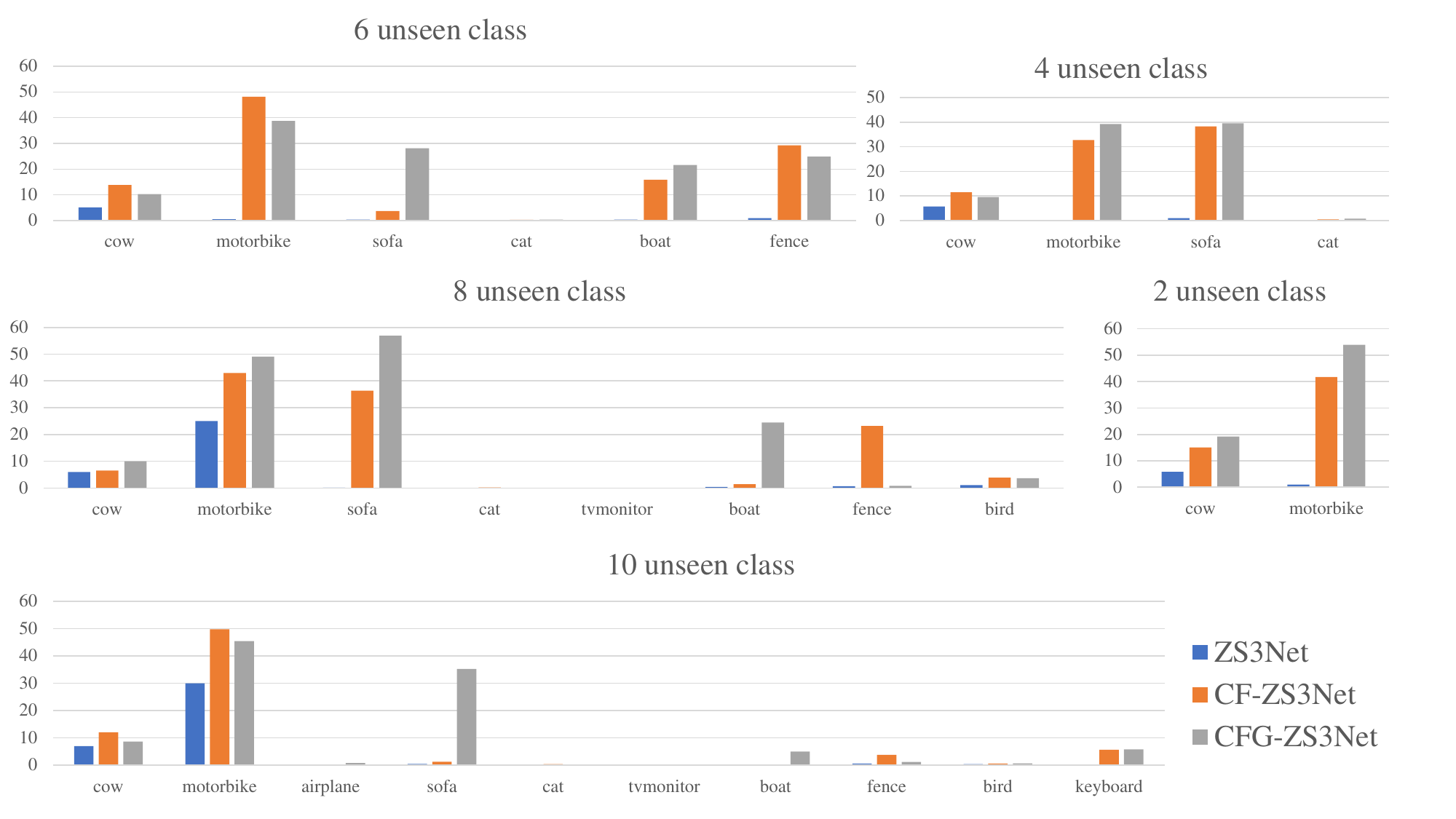}

    \caption{ Comparison of Counterfactual ZS3Net (CF-ZS3Net), CF-ZS3Net with GCN structure (CFG-ZS3Net), and original ZS3Net on Pascal Context dataset with different zero-shot settings. We list single-class mIoU performances in five experiments. The specific numerical data are shown in Table~\ref{tab:detail}} 

    \label{fig:table1}
\end{figure*}

\section{Experiments}
In this section, we evaluated the counterfactual generative training strategy. Compared with state-of-art models, the consistently better performance on different models and datasets demonstrates the correctness and feasibility of our strategy. The dataset, baselines, hyper-parameters, and metrics in the experiment are introduced. 

\subsection{Dataset}\label{sec:dataset}
We evaluate our counterfactual generative model on two challenging and sufficient datasets: Pascal-VOC 2012~\cite{pearl2009causal} and Pascal-Context~\cite{mottaghi2014role}. Both of them are widely used segmentation benchmarks containing indoor and outdoor images with corresponding annotations. 

\noindent \textbf{Pascal-VOC 2012} In our experiments, Our Pascal-VOC 2012 dataset contains 1464 images for training and 1449 images for validation. This dataset involves 20 object classes and one background class. In this work, the accuracy of background prediction is also accounted for in the overall performance. 

\noindent \textbf{Pascal Context} The Pascal Context provides full annotations of images in Pascal-VOC 2010. The meaning of full is that it not only gives the semantic segmentation of objects but also scene stuff such as the sky. The Pascal Context includes 4998 training and 5105 validation images. There are more than 400 classes in these images, following previous semantic segmentation works, we adopt the most frequent 59 classes of them to evaluate our models.

\subsection{Settings and Evaluations} 

We employ Word2Vec~\cite{mikolov2013distributed} trained on Google News as the word embedding model and DeeplabV3+~\cite{chen2018encoder} system adopting ResNet-101~\cite{he2016deep} to extract features. Different from previous work, we directly block the image signal of unseen classes rather than blocking pixel-level features because of the atrous convolution layers in DeepLabV3+: besides resolution enhancement, this convolution method also enlarges the view of the filter so the information of the unseen object can still influence the neighboring pixel-level features. The classes of objects we wipe out in certain experiments are listed in Table~\ref{tab:all}. Like previous weaker supervision work~\cite{bucher2019zero,toldo2020unsupervised}, the semantic annotations are available for all the training sets. In the testing period, either seen or unseen objects can appear in the image, namely ``generalized zero-shot”~\cite{chao2016empirical}. We adopt mean intersection-over-union (mIoU) and harmonic intersection-over-union (hIoU) for evaluation standards. The hIou can be calculated by mIou on seen and unseen classes:
\begin{equation}\label{eq:hiou}
hIoU=\frac{2 \cdot mIoU_{\text {seen }} \cdot mIoU_{\text {unseen }}}{mIoU_{\text {seen }}+m I o U_{\text {unseen }}}
\end{equation}

\subsection{Baseline} 
In this work, we adopt ZS3Net and SPNet as our baselines. Both of them are classical zero-shot semantic segmentation models. The SPNet also declares itself as a generative model but it has a different causal graph from causal structures of traditional generative zero-shot semantic segmentation like ZS3Net. While some other works surpass their performance with auxiliary structures, remember that we aim to alleviate the bias between seen and unseen classes. We just apply these two models as our baselines in order to get rid of noisy causal effects from other auxiliary structures.

\subsection{Implementation details} 
All classifiers in our work are Conv layers with $1\times1$ kernel size. We use the initial learning rate of $7\times10^{-3}$, the momentum of 0.9, and weight decay of $5\times10^{-4}$. The generator of fake features consists of three fully-connected layers and ReLU functions. The generator is trained by Adam optimizer with the learning rate $2\times10^{-4}$. The initial value of $a$ and $c_{1}+c_{2}$ is all zero. 
\begin{table}
\centering

\scalebox{0.85}{
\begin{tabular}{l cccc}
    \hline
    \toprule
    CFG-ZS3Net  & hIoU & overall mIoU &  Seen mIoU & Unseen mIoU\\
    \midrule
    2 unseen & \bf 38.21 & 38.96 & 39.01 & \bf 37.45 \\
    4 unseen & 25.64 & 29.63 & 30.15 & \bf 22.31 \\
    6 unseen & \bf 23.65 & 26.95 & 27.64 & \bf 20.68 \\
    8 unseen & \bf 21.89 & \bf 26.32 & 27.57 & \bf 18.16 \\
    10 unseen & \bf 13.1 & 16.8 & 18.1 & \bf 10.27 \\
    \bottomrule
    \hline
\end{tabular}
}
\vspace{2mm}
\caption{Supplementary to Table~\ref{tab:all}. Performance $(mIoU, hIoU)$ of CFG-ZS3Net on Pascal Context dataset. The unseen class setting is the same as Table~\ref{tab:all}. We {\bf highlight} the results that are higher than corresponding accuracy of CF-ZS3Net.}
\label{tab:detail}
\end{table}

\begin{figure}
\centering
{
\subfigure[]{
\label{fig:dis1}
\includegraphics[width=0.45\linewidth]{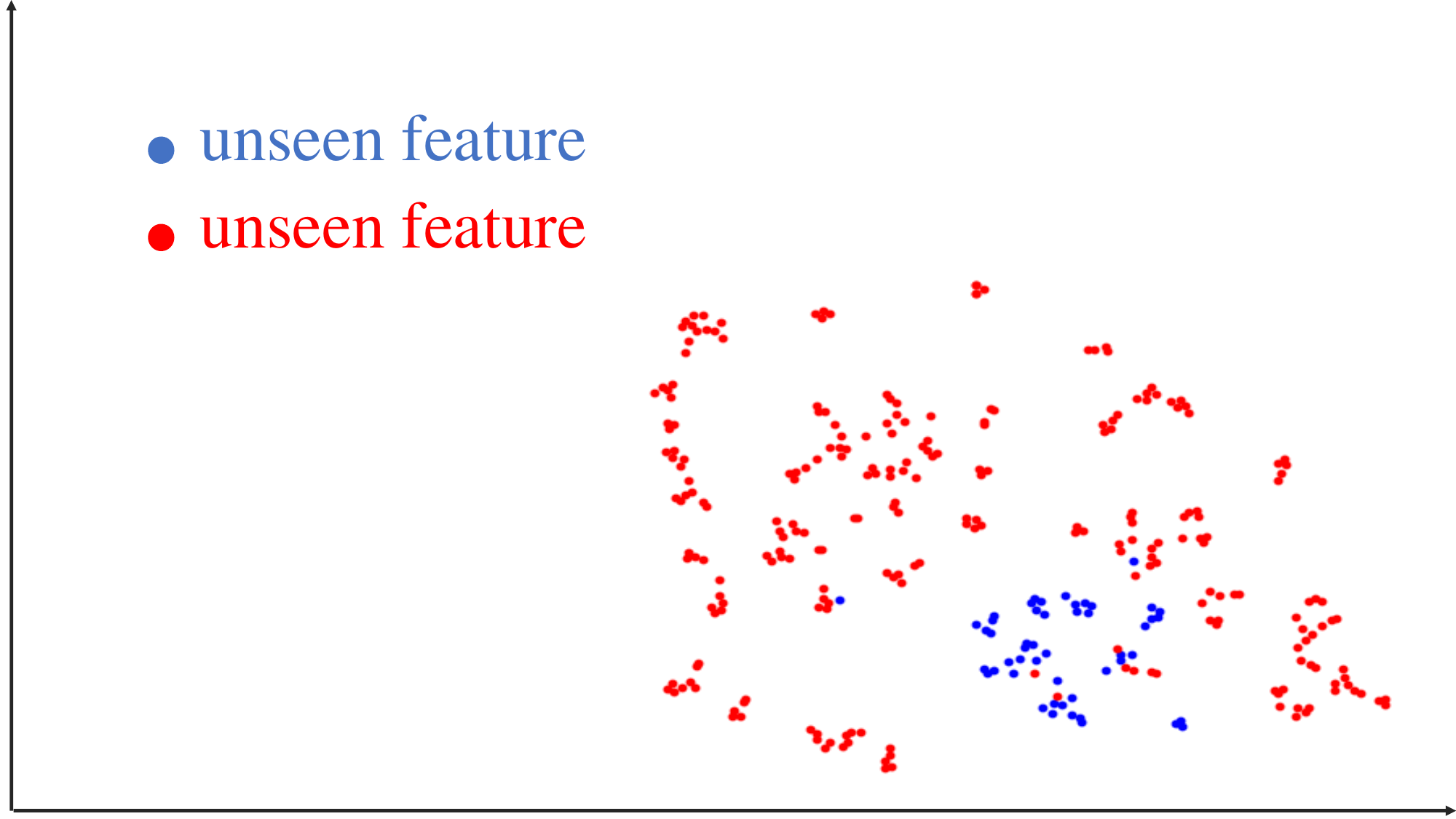}
}
\subfigure[]{
\label{fig:dis2}
\includegraphics[width=0.45\linewidth]{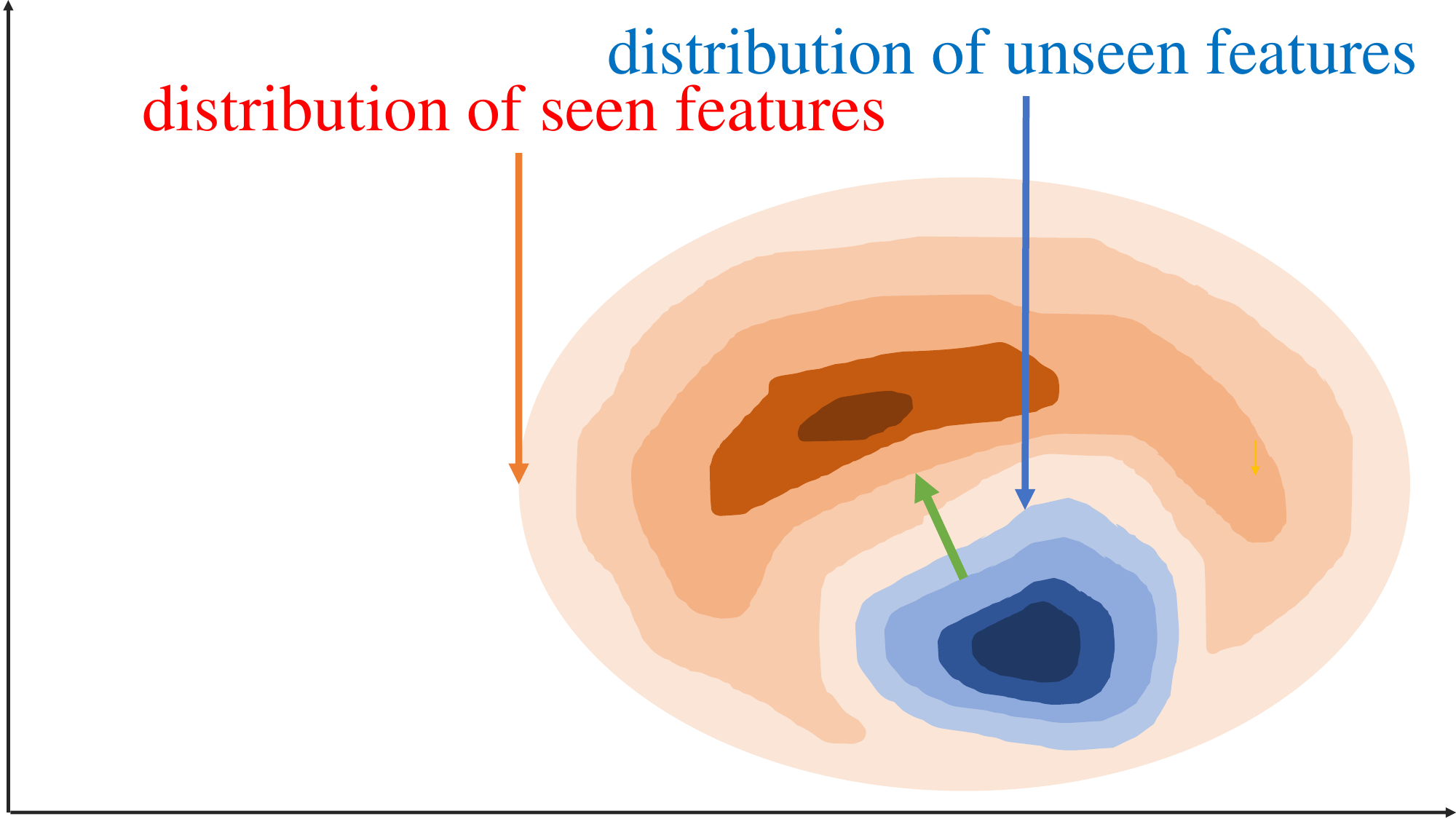}
}
\subfigure[]{
\label{fig:dis3}
\includegraphics[width=0.45\linewidth]{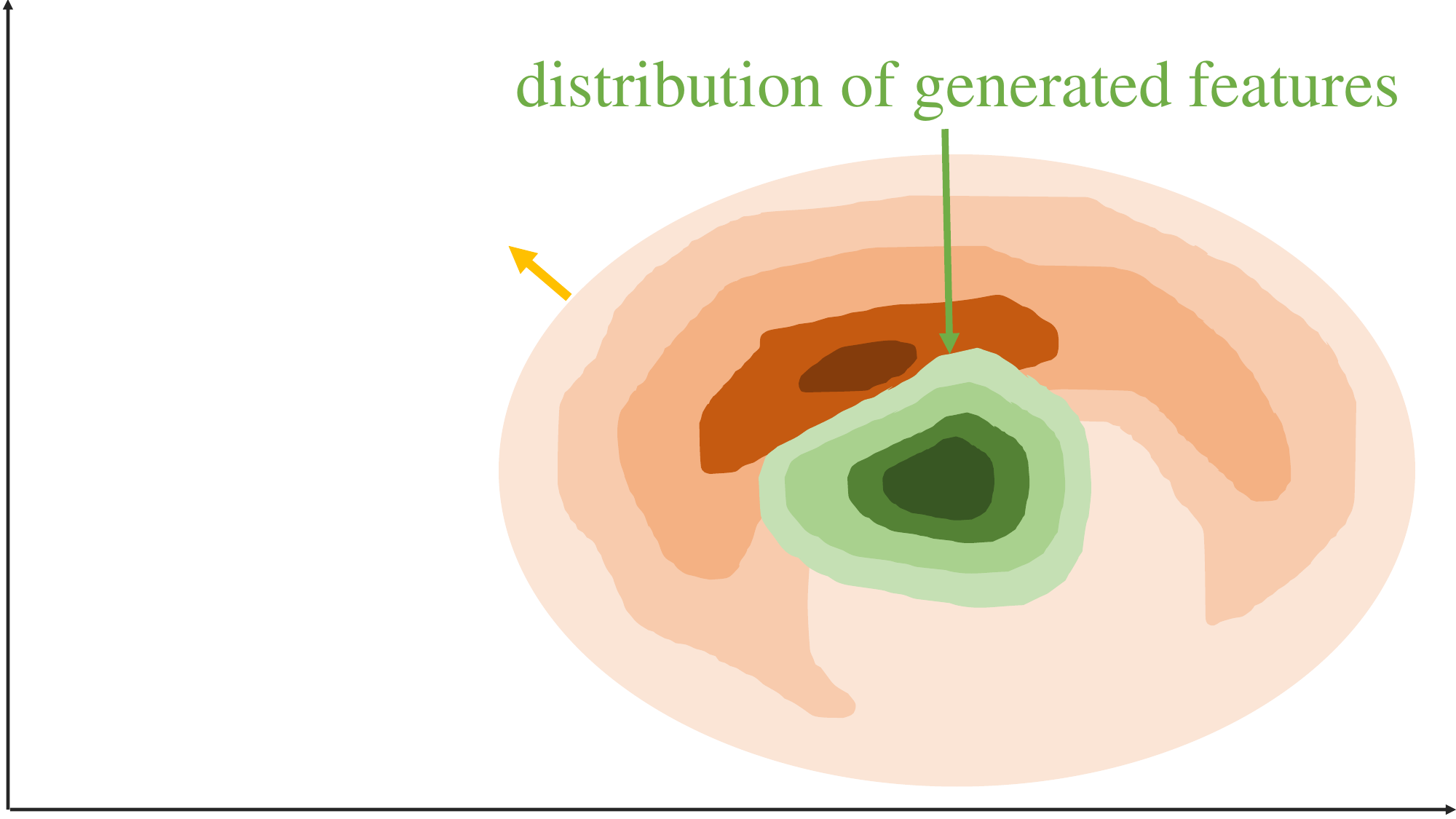}
}
\subfigure[]{
\label{fig:dis4}
\includegraphics[width=0.45\linewidth]{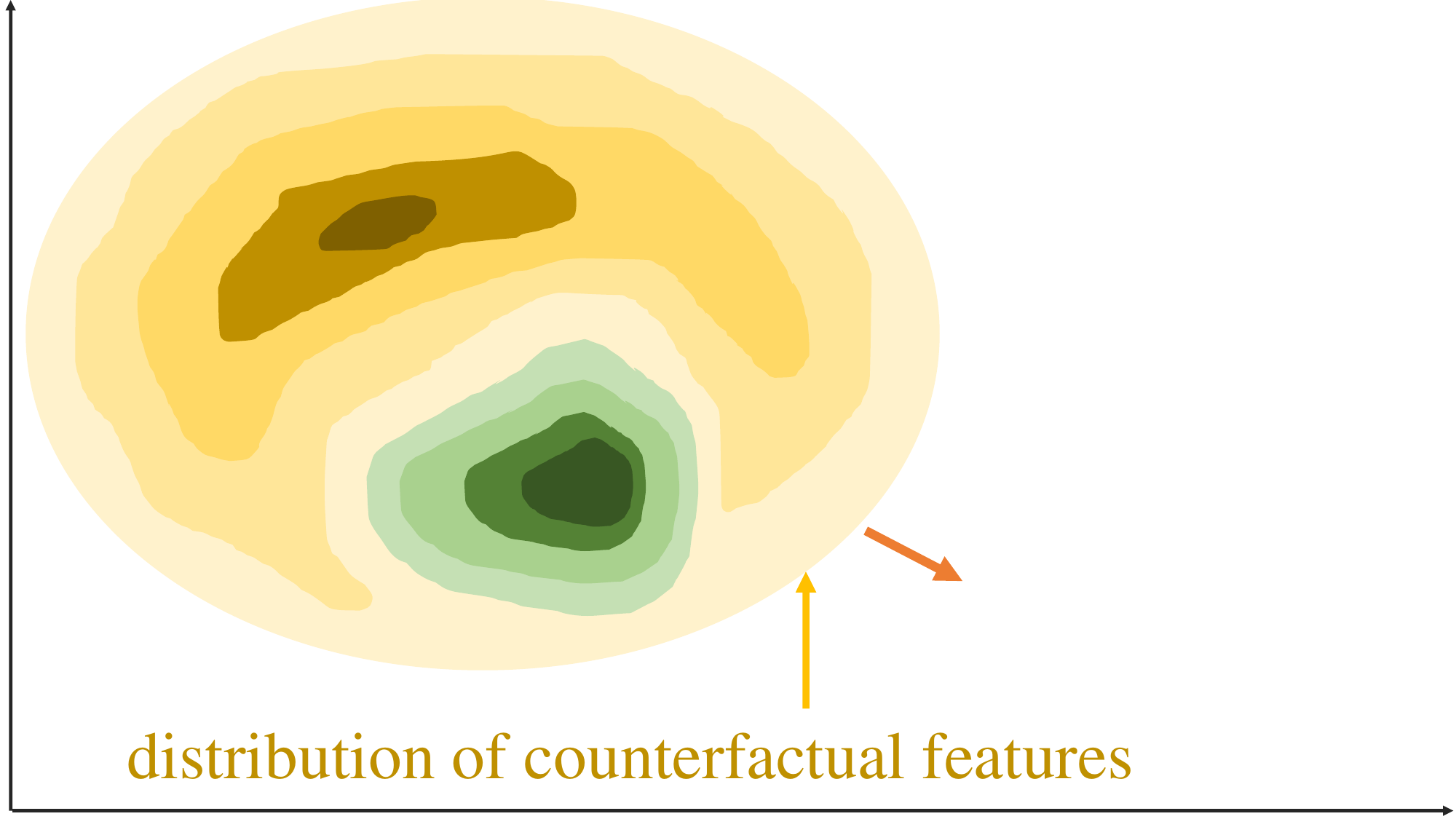}
}
}

\caption{(a) t-SNE~\cite{van2008visualizing} plot of feature sample from different class in Pascal-Context dataset. \textcolor{red}{Red spots} represent features of unseen classes and \textcolor{blue}{Blue spot} as those of seen classes. (b) The abstract distribution of seen class and unseen class. \textcolor{blue}{Blue parts} and \textcolor{red}{red parts} are correspondingly unseen and seen class distributions. Their borderline can be seen as the decision boundary of classifiers. \textcolor[RGB]{84,139,84}{Green arrow} means the offset trend of generated feature (c) \textcolor[RGB]{84,139,84}{Green part} is the guessing distribution of generated features. \textcolor[RGB]{255,215,0}{Yellow arrow} means the direction of our counterfactual features. (d)  \textcolor[RGB]{255,215,0}{Yellow part} is the distribution of counterfactual features. \textcolor{red}{Red arrow} means the offset trend of whole feature distribution.}
\label{fig:distribution}
\end{figure}

\subsection{Analysis}
In Table~\ref{tab:all} we compare ZS3net, SPNet with their counterfactual training vision in different zero-shot settings and datasets. 

\noindent \textbf{Pascal-VOC 2012} Actually, there are several causal zero-shot learning works before, some of them also take advantage of the counterfactual method on image recognition. Although their model indeed developed the accuracy of unseen class recognition, the accuracy of the seen classes is lower than the state-of-the-art models. In our experiments, while there are ten classes unseen to the training set, our model can still surpass the performance of the original ZS3Net on both seen and unseen classes. We can also discover that our method did not improve the performance of SPNet obviously. It also makes sense that in Section~\ref{sec:method}, we declare that parameters in SPNet have different causal-effect relationships. In ten class unseen situations, CF-ZS3Net improves the same setting state-of-art model by about $2\%\; mIoU$ on unseen classes and $31\%\;mIoU$ on seen classes. It represents that the generative fake feature is quite similar to the real feature of the unseen class so the classifier is confused with what is like the real feature of seen classes. Once this misunderstanding gets fixed, the accuracy of both seen and unseen classes get improved.

\noindent \textbf{Pascal Context} Compared to Pascal-VOC, Pascal-Context is a more challenging dataset. The length of the class category is three times more than Pascal-VOC even without unlabeled objects in images and the varieties of objects in one image are much higher than images in Pascal-VOC which commonly only include two or three objects. Ignoring the accuracy of a single class, we simply divide the classes into two groups: seen class and unseen class, and plot the feature distribution of these two classes. In the experiments, the feature of the unseen class is unavailable to the model so we use GAN to generate features of the unseen class to train the model. However, in this process, the distribution of generated features is unavoidable offset to the feature of the seen class like Figure~\ref{fig:dis2}. The direct consequence of this offsetting is the expansion of the misleading zone which comes from two parts of original seen and unseen feature distribution. The more large the seen class is the greater offset and the larger the misleading zone. 

Previous work~\cite{cui2019class,bucher2019zero} mainly did two things: getting the initially generated features closer to the real features of the unseen class or improving the weight of the unseen class in the classifier. The first method undoubtedly requires more advanced networks and resources including data and hardware. The second rough method can enlarge the distribution of generated features but will damage the accuracy of the seen class. Our method gives a different direction to solve the problem: we firstly send generated and real features to different classifiers to allow us to move them separately, then we create counterfactual features and computing suitable offsetting directions through deep networks. In Equation~\ref{eq:fl}, we move the generated features for convenience while in Figure~\ref{fig:dis3} , we move real features to flatter the definition of ``counterfactual’’: since we always regard the object of seen class as ``factual’’. After the distribution of generated features and counterfactual features get their relatively correct position, we perform another offset like Figure~\ref{fig:dis4}. This method allowed us to improve the accuracy of both seen and unseen classes under recent resources. In the larger dataset, this improvement is more obvious. In our experiments, our training strategy can improve $24\%\;mIoU$ on unseen classes and $19\%\;mIoU$ on seen classes. Remember that the Figure~\ref{fig:distribution} is just an abstract representation under ideal conditions. In the real application, there are more classes and every class may distribute among the group and may require more training parameters to simulate offset than our previous Equations which only pursue a globally optimal solution.

\begin{figure}
    \centering
    \includegraphics[width=0.95\linewidth]{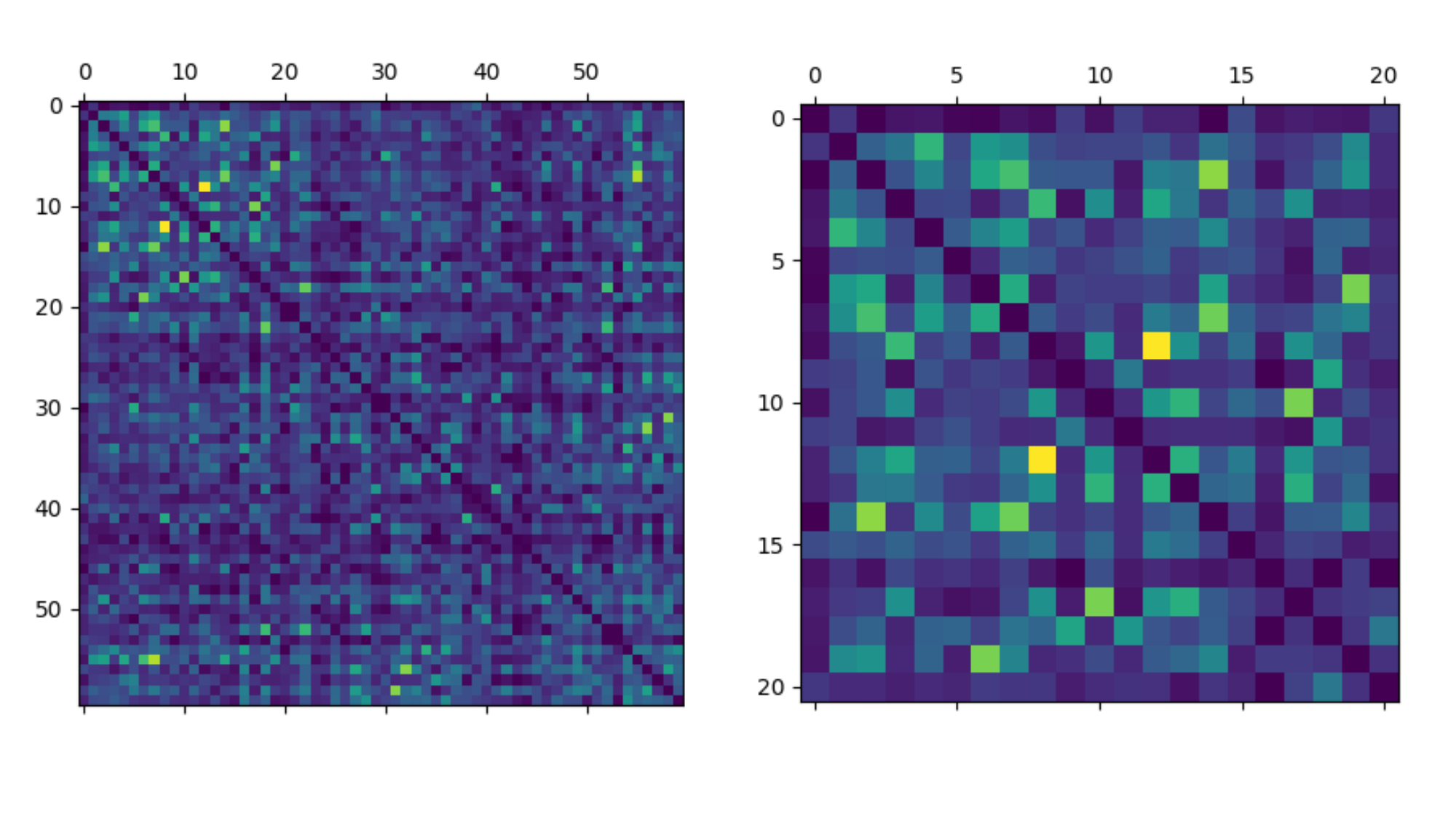}
    \caption{: Visualization of the generated adjacency matrices of Pascal-VOC 2012 and Pascal Context dataset. The darker cube means the higher similarity between the word embedding of corresponding classes} 
    \label{fig:adj}
\end{figure}

\noindent \textbf{GCN structure} Since our training strategy mainly focuses on the more macroscopic probability distribution, we can still make use of previous research to promise the generated features distribution to have a similar ``shape’’ like unseen features. In Table~\ref{fig:table1} we list concrete mIoU of each single unseen class. The overall mIoUs of our model with GCN structure are averaged $2\%\;mIoU$ higher than vision without this structure. As shown in Figure~\ref{fig:adj}, the number of classes in the Pascal-VOC dataset restricts the containing information of the generated adjacent matrix so we only show the results on Pascal Context which has more obvious improvement.  We can find in Figure that not every class in CFG-ZS3Net is higher than CF-ZS3Net. That is because not every class has their similar class in the dataset to learn how to generate corresponding fake features. Even if there exist high cosine similarities between word embedding of two classes did not necessarily mean their method to generate fake features is the same. So we may need some more flexible graph convolution methods to develop the performance. Our experiments also demonstrate that our training strategy can adopt auxiliary structures to generate fake features like previous works. It is not opposed to our feature distribution theory. Like what we said in the last sentence, in the real application the unseen feature is scattered and our training strategy only pursues the best global solution. If we want to develop some single class accuracy like Figure~\ref{fig:table1}, we need additional tools to help us. And these structures are hard to be applied to SPNet.


\section{Conclusion}
In this work, we propose a new training strategy on the generative zero-shot semantic segmentation task based on causal inference. We presented our own auxiliary structure to demonstrate that similar structures in previous also can be adopted by our framework. We apply our method on a state-of-the-art model: ZS3Net, and surpass its performance on Pascal-VOC 2012 and Pascal Context dataset. We alleviate the imbalance in the original model and explain the reason why we could improve the accuracy of both seen and unseen objects. There is also much work waiting for the future researcher. For instance, Siamese Network may able to get the zero-shot semantic segmentation task rid of the word embeddings by calculating similarity between classes. More advanced work in missing data imputation which has many similarities to zero-shot learning may be able to empower this task.

\bibliographystyle{unsrt}  
\bibliography{template}

\end{document}